\documentclass[11pt]{article}

\usepackage[preprint]{acl}
\usepackage{enumitem}

\usepackage{times}
\usepackage{latexsym}
\usepackage{booktabs}
\usepackage{multirow}
\usepackage{adjustbox}
\usepackage{amsmath,amssymb}
\usepackage[T1]{fontenc}
\usepackage{tcolorbox}
\tcbuselibrary{breakable}

\usepackage[utf8]{inputenc}

\usepackage{microtype}

\usepackage{inconsolata}

\usepackage{graphicx}

\title{Multimodal In-context Learning for ASR of Low-resource Languages}

\author{Zhaolin Li \and Jan Niehues \\
        Karlsruhe Institute of Technology \\Karlsruhe, Germany \\ \{zhaolin.li, jan.niehues\}@kit.edu}

\begin{document}
\maketitle
\begin{abstract}
Automatic speech recognition (ASR) still covers only a small fraction of the world’s languages, mainly due to supervised data scarcity. In-context learning (ICL) with large language models (LLMs) addresses this problem, but prior work largely focuses on high-resource languages covered during training and text-only settings. This paper investigates whether speech LLMs can learn unseen languages with multimodal ICL (MICL),  and how this learning can be used to improve ASR. We conduct experiments with two speech LLMs, Phi4 and Qwen3-Omni, on three diverse endangered languages. Firstly, we find that MICL is effective for unseen languages, leveraging both speech and text modalities. We further show that cross-lingual transfer learning improves MICL efficiency on target languages without training on them. Moreover, we analyze attention patterns to interpret MICL mechanisms, and we observe layer-dependent preferences between audio and text context, with an overall bias towards text. Finally, we show that prompt-based ASR with speech LLMs performs poorly on unseen languages, motivating a simple ASR system that combines a stronger acoustic model with a speech LLM via MICL-based selection of acoustic hypotheses. Results show that MICL consistently improves ASR performance, and that cross-lingual transfer learning matches or outperforms corpus-trained language models without using target-language data. Our code is publicly available\footnote{\url{https://github.com/ZL-KA/MICL}}.

\end{abstract}

\section{Introduction}

With more than 7,000 languages spoken worldwide, current Automatic Speech Recognition (ASR) models support only a small fraction of them, primarily due to the scarcity of labeled data across languages \cite{omnilingual2025omnilingual}. To mitigate this limitation, recent studies explore approaches such as self-supervised learning \cite{chen-etal-2024-towards-robust, pratap2024scaling, omnilingual2025omnilingual}, foundation model adaptation \cite{qian24_interspeech, fong25_interspeech, schmidt2025fleursslumassivelymultilingualbenchmark, li-niehues-2025-enhance}, and synthetic data augmentation \cite{bamfo-odoom-etal-2024-synthetic, li-etal-2025-kits}. Despite these advances, ASR performance remains suboptimal and lacks robustness, as existing methods continue to rely on limited amounts of high-quality annotated data that fail to capture sufficient diversity across domains, speakers, and recording conditions.

Recent advances in speech large language models (LLMs) offer a promising alternative. By leveraging large-scale multilingual and multitask data, speech LLMs can improve robustness through task understanding and reasoning \cite{chu2024qwen2, ghosh2025audio, ambilduke-etal-2025-tower}. However, their performance is still largely limited to high-resource languages, since the majority of training data is dominated by those languages.

For low-resource languages in particular, in-context learning (ICL) has attracted attention as it enables models to adapt using only a small number of demonstration examples provided in the prompt. Prior studies show that linguistic descriptions and example samples can effectively teach languages to LLMs \cite{zhang-etal-2024-hire, zhang-etal-2024-teaching, li25ca_interspeech, li-etal-2025-context-learning, ma-etal-2025-exploring-role, pei-etal-2025-understanding}. Nevertheless, most existing ICL research for low-resource languages focuses on the text modality and text-based LLMs, leaving speech largely underexplored. Recent work on multimodal in-context learning (MICL) shows that paired audio–text demonstrations can improve ASR robustness, including speaker, accent, and domain adaptation \cite{roll2025context, zheng2025ticl}. However, they mainly focus on languages already covered by the target models. 

The work most closely related to ours is \cite{hsu2024meta}. While it studies similar data and task settings, its focus is primarily on Whisper-style speech foundation models rather than speech LLMs. This distinction matters because speech LLMs differ from such models in both scope and capability, particularly in their ability to process longer audio context and follow broader natural-language instructions \cite{radford2023robust}. Therefore, the findings of \cite{hsu2024meta} are not directly sufficient for understanding our setting, which motivates the present study.

In this work, we investigate how MICL can address data scarcity in building ASR systems for low-resource languages. We first study whether MICL enables speech LLMs to learn languages from provided samples. Using three linguistically diverse endangered languages, we evaluate two state-of-the-art open-source speech LLMs, Phi4 \cite{abouelenin2025phi} and Qwen3-Omni \cite{xu2025qwen3omnitechnicalreport}. We then fine-tune the speech LLMs with instruction data in other languages and test whether the cross-lingual transfer learning commonly reported for traditional speech foundation models also appear in instruction-tuned speech LLMs.  \cite{conneau21_interspeech, ma-etal-2025-cross}. We also analyze how MICL is internally realized within the model to interpret its underlying attention mechanisms. Finally, we propose a simple ASR system that integrates an acoustic model with a speech LLM through hypothesis selection, combining the strong recognition accuracy of acoustic models with the contextual learning capabilities of speech LLMs \cite{chen2025ml, geng2025evaluating}.

Our findings are summarized as follows:
\begin{itemize}[itemsep=0pt, topsep=0pt]
    \item MICL enables speech LLMs to learn uncovered languages, benefiting from both speech and text modalities (\S \ref{sec:MICL}).
    \item Cross-lingual transfer learning improves language learning efficiency (\S \ref{sec:cross-lingual}).
    \item Speech LLMs exhibit layer-dependent attention preferences for audio versus text samples, and overall they allocate more attention to text than to audio (\S \ref{sec:MICL_interpret}).
    \item MICL enhances ASR performance and that cross-lingual transfer learning can match or outperform corpus-trained language models even without target-language data.
    (\S \ref{sec:hypo_selection}).
\end{itemize}

\section{Approach}

% Multi-modal In-context Language Learning

Following the interleaved-modality design of speech LLMs, we study MICL for speech recognition \cite{abouelenin2025phi, xu2025qwen3omnitechnicalreport, nguyen2025spirit}. We first formalize the MICL task in Section \ref{sec:task}. We then introduce different prompt designs to analyze the impact of modality composition in MICL, as described in Section \ref{sec:prompt_design}. To ensure effective in-context demonstrations, we implement several sample selection strategies, detailed in Section \ref{sec:sample_select}. In addition, we explore cross-lingual instruction fine-tuning to enhance MICL performance using multilingual resources that exclude the target languages in Section \ref{sec:XFT_approach}. Finally, we build an ASR hypothesis selection system that leverages MICL to improve acoustic-based ASR models in Section \ref{sec:hypo_select_approach}.

\subsection{Task Description} \label{sec:task}

We study in-context learning for ASR, where a model predicts a target transcription by conditioning on a set of in-context demonstrations. Let the demonstration set be
\begin{equation}
\mathcal{C} = \{c_i\}_{i=1}^N,
\end{equation}
where each $c_i$ denotes one demonstration. The exact content of each demonstration depends on the prompt design and is specified in Section~\ref{sec:prompt_design}.

Given a demonstration set $\mathcal{C}$ and a target audio input $a^*$, the model is asked to predict the corresponding transcription $t^*$. The multimodal model $f_\theta$, parameterized by $\theta$, defines a conditional probability distribution over the target transcription as
\begin{equation}
p_\theta(t^* \mid a^*, \mathcal{C}) = \prod_{j=1}^{T} p_\theta(w_j^* \mid w_{<j}^*, a^*, \mathcal{C}),
\label{eq:conditional_prob}
\end{equation}
where $w_{<j}^* = (w_1^*, \ldots, w_{j-1}^*)$ denotes the previously generated tokens and $T$ is the length of the target sequence.

This formulation provides a unified view of all settings considered in this work. Different prompt designs instantiate $\mathcal{C}$ with different modality combinations and provide either the presence or absence of the target audio $a^*$, allowing us to isolate how textual and acoustic context affects ASR performance.

\subsection{Prompt Modality Design} \label{sec:prompt_design}

To isolate the contribution of each modality, we consider three in-context learning settings that differ only in the information included in the demonstrations\footnote{Examples in Appendix~\ref{sec:app_prompt}}. For reference, we also include a standard ASR prompt corresponding to the task formulation commonly used for speech LLMs.

\textbf{T-ICL}. Each demonstration contains only text, i.e., $c_i = t_i$. This setting measures how much improvement can be obtained from textual context alone, such as lexical or orthographic patterns, without acoustic evidence \cite{hsu2024meta, li25ca_interspeech}.

\textbf{ICL}. Each demonstration also contains only text, i.e., $c_i = t_i$, while the model additionally receives the target audio $a^*$. This setting evaluates whether text-only demonstrations can guide transcription of an audio query. Comparing ICL with T-ICL isolates the contribution of the target audio.

\textbf{MICL}. Each demonstration contains paired audio and text, i.e., $c_i = (a_i, t_i)$. Together with the target audio $a^*$, this setting tests whether paired multimodal demonstrations provide additional benefit beyond text-only context.

\textbf{ASR}. This baseline does not use in-context demonstrations. The model directly transcribes the input audio $a^*$, representing the standard speech LLM inference setup.

\subsection{Sample Selection Strategy}\label{sec:sample_select}

Selecting relevant demonstrations is important for effective text-based ICL, especially when the target language is low-resource or unseen \cite{cahyawijaya-etal-2024-llms, li25ca_interspeech}. We therefore include a retrieval-based sample selection module in this MICL work. Following \cite{li25ca_interspeech}, we use SONAR \cite{Duquenne:2023:sonar_arxiv}, a multilingual and multimodal embedding model, to embed the candidate samples and the predicted transcript of target sample. We rank candidates samples by calculating cosine similarity to the target in the SONAR embedding space and select the top-$N$ samples as the in-context demonstrations. More details are provided in Appendix \ref{sec:app_sample_selection}.

\subsection{Language-specific and Cross-lingual Fine-tuning} \label{sec:XFT_approach}

In addition to fine-tuning on data from the target language, we explore cross-lingual instruction fine-tuning to examine whether supervised adaptation on MICL tasks can improve generalization to unseen languages and enhance the model’s ability to follow MICL prompt formats. Accordingly, we investigate three fine-tuning strategies in this work:

\textbf{ASR-FT.} Standard ASR fine-tuning, where the model predicts the transcript from the target audio without in-context demonstrations.

\textbf{TFT.} Target-language fine-tuning, where the model is fine-tuned on ASR instances constructed from the target language. Each training instance contains a target audio utterance together with in-context demonstrations from the same language, and the model predicts the transcript of the target audio.

\textbf{XFT.} Cross-lingual fine-tuning, where the model is fine-tuned on ASR instances from multiple languages other than the evaluation language. In this setting, the fine-tuning data excludes the evaluation language entirely. Instead, training batches contain samples from a diverse set of auxiliary languages, each with its own in-context demonstrations and target audio. This setting is designed to examine whether multilingual transfer improves performance on unseen languages.

\subsection{ASR Hypothesis Selection System} \label{sec:hypo_select_approach}

We further develop an ASR hypothesis selection system that leverages MICL to re-rank hypotheses generated by an external acoustic-based ASR model, which is typically more robust than speech LLMs in low-resource settings \cite{chen2025ml, geng2025evaluating}. We also verify this in Appendix~\ref{sec:app_ft_asr}.

Given an input utterance $a_{1:S}$, an external ASR system first produces an $N$-best list of candidate transcriptions $\mathcal{H}=\{h^{(k)}\}_{k=1}^{N}$. We then score each hypothesis using two signals: (1) the acoustic score from the external ASR model, and (2) the LM score from a language model, which measures the contextual understanding. The final output is selected by re-ranking the $N$ candidates with a combined score:
\begin{align}
\hat{h} = \arg\max_{h^{(k)} \in \mathcal{H}} \Big[
\mathrm{Acoustic\_score}(h^{(k)}; a_{1:S}) \\
+ \mathrm{LM\_score}_{\mathrm{MICL}}(h^{(k)}; a_{1:S}, \mathcal{C})
\Big].
\end{align}
We compute $\mathrm{LM\_score}$ as the log-likelihood of the hypothesis tokens under the language model. In preliminary experiments, using the LM score alone performs worse than the combined approach; so we use the combined scoring through experiments. Implementation details are referred to in the scripts.

\section{Experimental setup}

\subsection{Datasets}

Because speech LLMs are often trained on large and diverse data sources, they may be exposed to more languages than than are documented during training. To investigate model performance on unseen languages, we conduct experiments on three endangered languages that are more likely to be uncovered during training, as shown in Table \ref{tab:dataset}: Khinalug (ISO 639-3: kjj, Northeast Caucasian ) \cite{li-etal-2024-speech}, Kichwa (ISO 639-3: que,  Quechuan ) \cite{taguchi-etal-2024-killkan}, Mboshi (ISO 639-3: mdw, Bantu ZoneC) \cite{godard-etal-2018-low}.  These three languages come from different language families with different source type, ensuring diversity in our experiments. Following other work, we lowercase all text and remove the punctuation in preprocessing.

We further investigate whether cross-lingual instruction fine-tuning improves performance on these evaluation languages. Specifically, we fine-tune the model on a multilingual MICL instruction dataset covering 143 languages derived from ML-SUPERB 2.0 \cite{shi24g_interspeech}. None of these languages overlap with our three evaluation languages, allowing us to test whether cross-lingual instruction fine-tuning improves generalization to unseen target languages.

\begin{table}[h]
\centering
\small
\begin{tabular}{cccccc}
\hline
Language    & Audio source                         & Train (h) & Dev+Test (h) \\ \hline
Khinalug  & Spontaneous                        & 2.14        & 0.49      \\
Kichwa               & Radio                              & 3.05        & 0.77    \\
Mboshi           & Reading                             & 3.93        & 0.53    \\  \hline
\end{tabular}
\caption{Dataset descriptive statistic.}
\label{tab:dataset}
\end{table}

\subsection{Models}

We experiment with two state-of-the-art open-source multimodal LLMs, Phi4 and Qwen3-Omni, which are trained on diverse instruction-following datasets covering multiple modalities. This makes them particularly well suited for MICL, where models must understand the task from provided examples and generalize to unseen inputs. Details regarding model versions and inference settings are provided in Appendix~\ref{sec:app:exp_setup}.

\subsection{Fine-tuning}

During fine-tuning, the training loss is computed only on the target transcription tokens, while the in-context audio--text demonstration pairs are treated as conditioning context. This objective encourages accurate transcription of the target audio while allowing the model to learn how to utilize the multimodal context provided by the MICL prompt \cite{abouelenin2025phi}. Under computational constraints, and based on our preliminary analysis of the number of in-context samples (Appendix~\ref{sec:app_ft_num_sample}), TFT and XFT experiments use training instances whose number of in-context samples is randomly selected from 1 to 10. Detailed hyperparameter settings are provided in Appendix~\ref{sec:app:exp_setup}.

Based on our preliminary comparison of fine-tuning strategies (Appendix~\ref{sec:app:ft_startegies}), we adopt LoRA on the decoder for fine-tuning and freeze the remaining model parameters, as this gives comparable performance to updating additional modules while being more parameter-efficient.

\subsection{Hypotheses Generation}

We use the MMS model \cite{pratap2024scaling} to generate recognition hypotheses, as it demonstrates strong performance in low-resource speech recognition due to its self-supervised learning architecture and broad language coverage. Specifically, we fine-tune the model on each target language and decode with beam search to obtain 10 hypotheses.

\subsection{Evaluation Metrics}

We evaluate ICL performance with perplexity. For each test instance, we construct the prompt input, append the gold transcription, and run the model in a teacher-forcing manner. Perplexity is computed only over the gold transcription tokens, with the prompt treated as conditioning context. Lower perplexity indicates that the model assigns higher probability to the correct transcription and thus exhibits greater confidence under the given prompt.

To evaluate ASR performance, we report word error rate (WER), where lower WER indicates better recognition accuracy. 

Because evaluating WER for all MICL configurations is computationally expensive, we use perplexity to evaluate the full configuration space and run ASR evaluation with WER only on selected settings. Therefore, we do not assume a direct analytical mapping between perplexity and WER. Instead, we use perplexity as a configuration-selection signal and later provide empirical evidence that lower perplexity is generally associated with lower WER in the matched systems evaluated in this work.

\begin{table*}[]
\centering
\small
\setlength{\tabcolsep}{4pt}
\begin{tabular}{ccccccccccc}
\hline
Language & Task  & 0 & 1 & 2 & 3 & 5 & 10 & 25 & 50 & 100 \\
\hline

% Khinalug & Qwen2.5-omni & ICL & Text samples &3075 & 507 & 258 & 186 & 96 & 82 & 66 & 72 & 72 \\

%  &  & ICL & Text samples \& target audio & 1159 & 658 & 444 & 401 & 133 & 108 & 87 & 79 & 74 \\
%   &  & ICL & Pair samples \& target audio  & 1228 & 513 & 312 & 358 & 129 & 107 & 87 & 79 & 78 \\
% &  & ASR &  -       & 2498  &      &     &     &     &     &     &     &     \\
% \cline{2-13}

Khinalug & T-ICL & 1302 & 289 & 146 & 201 & 69 & 57 & 40 & 44 & 43 \\

 &  ICL  & 54 & 28 & 17 & 14 & 11 & 10 & 9 & 11 & 15 \\
  & MICL  & 58 & 30 & 13 & 13 & 9 & 10 & 8 & 8 & 13 \\
  \cline{2-11}
 & ASR & \multicolumn{9}{c}{80}  \\

\hline

% Kichwa & Qwen2.5-omni & ICL & Text samples &6408 & 551 & 431 & 309 & 147 & 108 & 228 & 101 & 73 \\

%  &  & ICL & Text samples \& target audio & 4005 & 805 & 491 & 360 & 214 & 152 & 226 & 111 & 89 \\
%   &  & ICL & Pair samples \& target audio  & 4132 & 718 & 472 & 329 & 185 & 123 & 138 & 95 & 82 \\
% &  & ASR &  -       & 6405  &      &     &     &     &     &     &     &     \\
% \cline{2-13}

Kichwa & T-ICL &4292 & 417 & 184 & 170 & 101 & 82 & 652 & 153 & 41 \\
 &   ICL & 18 & 10 & 8 & 5 & 5 & 4 & 3 & 3 & 3 \\
  &   MICL & 17 & 7 & 6 & 5 & 4 & 4 & 3 & 3 & 4 \\
\cline{2-11}
& ASR &  \multicolumn{9}{c}{24}  \\

\hline

% Mboshi & Qwen2.5-omni & ICL & Text samples &5326 & 218 & 132 & 104 & 75 & 51 & 36 & 31 & 28 \\

%  &  & ICL & Text samples \& target audio &2439 & 324 & 193 & 154 & 112 & 77 & 52 & 42 & 36 \\
%   &  & ICL & Pair samples \& target audio  & 3038 & 402 & 185 & 140 & 100 & 68 & 47 & 39 & 35 \\
% &  & ASR &  -       & 6565  &      &     &     &     &     &     &     &     \\
% \cline{2-13}

Mboshi & T-ICL  & 2320 & 172 & 101 & 80 & 59 & 40 & 28 & 23 & 20 \\
 &  ICL & 178 & 51 & 37 & 29 & 21 & 16 & 11 & 10 & 9 \\
  & MICL & 189 & 34 & 24 & 17 & 13 & 10 & 7 & 7 & 9 \\
\cline{2-11}
&  ASR &     \multicolumn{9}{c}{242}    \\

\hline

\end{tabular}
\caption{Perplexity results for in-context learning with Qwen3-Omni. Numeric columns indicate the number of in-context samples. Task types follow the prompt designs described in Section~\ref{sec:prompt_design}.}
\label{tab:ppl_qwen}
\end{table*}

\section{Results and Analysis}

\subsection{Speech LLMs perform MICL}\label{sec:MICL}

Tables~\ref{tab:ppl_qwen} and~\ref{tab:ppl_Phi4} show results of Qwen3-Omni and Phi4. Because Phi4 and Qwen3-Omni use different tokenization and vocabulary sizes, absolute perplexity values are not directly comparable across models. We therefore focus on relative perplexity changes within each model under different prompting settings.

\textbf{Scaling behavior:} Looking at the results of the pretrained models, we observe a consistent pattern across all ICL settings: increasing the number of in-context samples generally lowers perplexity, even with long prompts containing up to 100 samples, demonstrating that in-context learning remains effective under long-context conditions.\footnote{The abnormally high perplexity for Khinalug with 3 samples is consistent across different settings (Appendix~\ref{sec:app_sample_selection}), indicating imperfect sample selection .}. Moreover, ICL prompts yield lower perplexity than the no-context ASR setting, indicating that both speech LLMs can perform ICL by exploiting demonstrations to better predict target transcriptions for languages not seen during training.

\textbf{Modality utilization:} When comparing T-ICL and ICL for the pretrained models, we find that incorporating the target audio consistently leads to lower perplexity. This suggests that the models effectively leverage the audio modality and understand the ASR task beyond pure text prompting.

Comparing ICL and MICL for the pre-trained models, Qwen3-Omni and Phi4 show different behaviors. For Qwen3-Omni in Table~\ref{tab:ppl_qwen}, MICL consistently outperform ICL even with up to 100 samples, suggesting that Qwen3-Omni benefits from in-context audio across the full context range we evaluate. In contrast,  for Phi4 in Table~\ref{tab:ppl_Phi4}, MICL tends to provide larger gains when the number of in-context samples is small. In particular, MICL consistently outperforms ICL when the context contains at most three samples.

We hypothesize that these differences stem from the models’ training strategies, particularly their audio training scale, multilingual audio coverage, and long-context audio modeling capacity. Phi4 is trained on 2.3M hours of audio data and supports eight languages, whereas Qwen3-Omni is trained with 20M hours of audio data. It supports speech understanding in 19 languages and can process audio recordings of up to 40 minutes per instance \cite{abouelenin2025phi, xu2025qwen3omnitechnicalreport}. These design differences may help explain why Qwen3-Omni more consistently leverages audios under long-context MICL, while Phi4 shows clearer benefits primarily in shorter-context settings.

\begin{table*}[]
\centering
\small
\setlength{\tabcolsep}{4pt}
\begin{tabular}{cccccccccccc}
\hline
Language & Model & Task  & 0 & 1 & 2 & 3 & 5 & 10 & 25 & 50 & 100 \\
\hline

Khinalug & Phi4 & T-ICL & 7435 & 1347 & 955 & 779 & 328 & 248 & 193 & 166 & 175 \\
 &  & ICL     & 1445 & 909 & 606 & 1778 & 189 & 154 & 143 & 147 & 158 \\
 &  & MICL   & 1508 & 485 & 318 & 545 & 261 & 230 & 273 & 284 & 241 \\
 \cline{3-12}
 &  & ASR &   \multicolumn{9}{c}{2045}   \\
\cline{2-12}
 &  Phi4 XFT & T-ICL   & 3282 & 943 & 792 & 661 & 350 & 266 & 197 & 176 & 185 \\
 &  & ICL     & 771 & 235 & 291 & 269 & 73 & 67 & 65 & 70 & 78 \\
 &  & MICL    & 838 & 188 & 193 & 210 & 73 & 68 & 76 & 95 & 98 \\
\cline{2-12}
 & Phi4 TFT & ICL  &  148 & 38 & 19 & 25 & 15 & 15 & 14 & 15 & 16 \\
& & MICL     &231 & 63 & 53 & 70 & 27 & 25 & 38 & 55 & 72 \\
% \cline{3-12}
%   &  & ASR &    \multicolumn{9}{c}{26}     \\
\hline

Kichwa & Phi4 & T-ICL &35157 & 3141 & 1568 & 1125 & 836 & 424 & 534 & 266 & 253 \\
 &  & ICL  & 132   & 65  & 78  & 70  & 74  & 39 & 30 & 33 & 26 \\
  &  & MICL   & 36   & 64  & 58  & 43  & 48  & 31 & 40 & 58 & 69 \\
\cline{3-12}
&  & ASR     &  \multicolumn{9}{c}{210}  \\
\cline{2-12}
& Phi4 XFT & T-ICL  &8287 & 1352 & 795 & 551 & 357 & 236 & 241 & 156 & 156 \\
 &  & ICL  & 69 & 22 & 16 & 13 & 11 & 10 & 8 & 8 & 8 \\
  &  & MICL  & 65 & 22 & 19 & 14 & 13 & 11 & 10 & 10 & 14 \\
\cline{2-12}
  & Phi4 TFT & ICL  &5 & 5 & 7 & 8 & 7 & 7 & 7 & 7 & 6 \\
&  & MICL  & 11 & 8 & 8 & 7 & 6 & 5 & 6 & 6 & 12 \\
% \cline{3-12}
%   &  & ASR &    \multicolumn{9}{c}{5}     \\
\hline

Mboshi & Phi4 & T-ICL  &11407 & 436 & 240 & 190 & 131 & 84 & 62 & 54 & 52\\
 &  & ICL  & 1125  & 232 & 159 & 126 & 97  & 71 & 51 & 42 & 40 \\
  &  & MICL   & 1192  & 249 & 161 & 116 & 93  & 75 & 63 & 70 & 84 \\
  \cline{3-12}
&  & ASR     &  \multicolumn{9}{c}{3433}  \\
\cline{2-12}

& Phi4 XFT & T-ICL  & 4897 & 367 & 217 & 171 & 126 & 85 & 63 & 58 & 57  \\
 &  & ICL & 781 & 116 & 71 & 56 & 44 & 33 & 26 & 24 & 25  \\
  &  & MICL  & 609 & 83 & 55 & 43 & 36 & 28 & 26 & 30 & 30 \\
\cline{2-12}

 &  Phi4 TFT  & ICL  & 7 & 8 & 7 & 7 & 7 & 7 & 6 & 6 & 6 \\
 & & MICL   & 11 & 12 & 12 & 14 & 16 & 16 & 14 & 13 & 17 \\
 % \cline{3-12}
 %  &  & ASR &    \multicolumn{9}{c}{7} \\

\hline

\end{tabular}
\caption{Perplexity results for in-context learning with Phi4. Numeric columns indicate the number of in-context samples. Task types follow the prompt designs described in Section~\ref{sec:prompt_design}. Phi4 denotes the pre-trained model, while XFT indicates cross-lingual fine-tuned models, and TFT indicates target-language fine-tuned models with the in-context samples using the supervised data of the language.}
\label{tab:ppl_Phi4}
\end{table*}

\subsection{Cross-lingual fine-tuning benefits uncovered languages}\label{sec:cross-lingual}

We explore cross-lingual transfer learning to leverage multilingual resources while excluding the target languages, as explained in Section~\ref{sec:XFT_approach}. Due to computational limitations, we conduct experiments with Phi4 only. As shown in Table~\ref{tab:ppl_Phi4}, fine-tuning consistently improves Phi4 across all evaluated unseen languages. Notably, for Kichwa, cross-lingual fine-tuning achieves performance close to that of target-language fine-tuning, despite Kichwa being excluded from the cross-lingual fine-tuning data. This result indicates that increasing language diversity during model adaptation can enhance the generalization ability of speech LLMs, including transfer to languages not observed during fine-tuning.

We further analyze how cross-lingual instruction fine-tuning interacts with MICL. Consistent with the findings in Section~\ref{sec:MICL}, Phi4 benefits more from paired audio–text in-context demonstrations, particularly in low-shot settings. After instruction fine-tuning, the model follows the MICL prompt format more reliably and utilizes contextual information more effectively; however, this improvement is observed only with up to ten demonstrations, matching the maximum number of samples used during fine-tuning. Given that our fine-tuning data is limited in both size and diversity, and that the model is fine-tuned with a restricted range of in-context sample counts, we assume that scaling up instruction fine-tuning—both in data volume and diversity—could further improve speech context understanding and enhance the effectiveness of MICL.

\subsection{Interpretable MICL} \label{sec:MICL_interpret}
We analyze how speech LLMs utilize multimodal demonstrations in MICL by examining attention weights. Our analysis focuses on Phi4 as a representative speech LLM, in which a decoder attends over interleaved text and speech representations produced by modality-specific encoders.
% While architectural and training differences may affect quantitative patterns, the qualitative phenomena we report are expected to generalize to models with similar decoder-centric multimodal integration.
We note that attention is not a perfect explanation of model behavior. Speech LLMs can exhibit attention artifacts, such as attention sinks, and strong head-level specialization. Therefore, we treat attention statistics as a diagnostic signal rather than causal evidence \cite{chefer2021transformer,kang2025see,kang2025your}. 

We experiment on the Khinalug dataset. For each example, we compute the attention mass from generated output tokens to the tokens belonging to prompt, demonstration audios and texts, target audio and ASR task. We only report values on demonstrations for analysis in this section, and present the entire scores in Table \ref{tab:attn_allocation} in Appendix~\ref{sec:app_attn_allocation} .

\textbf{Imbalanced attention on modalities:} We first quantify how much the model attends to the audio part versus the text part among all attention goes into the in-context demonstrations. Here, we average the attention mass across heads and layers, and then across the evaluation set. Table~\ref{tab:attn_ratio_audio_text} that the model allocates more attention to text tokens than to audio tokens in the demonstrations, even though the audio representation is roughly three times longer than the text representation in the Phi4 setting (Appendix~\ref{sec_app_audio_text_repre_lengths}). Referring to previous work about vision-based LLMs \cite{baldassini2024makes, chentrue} showing multimodal LLMs mainly learn from text context and needs to pay more attention on visual modality, our finding confirms a similar imbalanced attention distribution for the speech modality. This observation indicates the potential to improve speech processing performance by designing LLMs to pay more attention to audio.

We further observe that increasing the number of in-context samples shifts relatively more attention toward the audio samples. A possible explanation is that LLM is capable of understanding the task by learning the correspondence between audio and text within each example pair, and thus tends to increase attention on audios, which are as important as texts. We also show this capability of understanding the correspondence within one pair in Section~\ref{sec:ablation_attention}.

\begin{table}[h]
    \centering
    \small
    \begin{tabular}{ccc} \hline
    \# Samples     & Audios & Texts  \\ \hline
    % 1     & 37.2\% & 62.8\% \\

    % 3 & 39.6\%	& 60.4\%\\
    % 5 & 41.0\% & 59.0\% \\
    % 10 & 42.9\% & 57.1\% \\
 1 &   30.5\%  &	69.5\% \\
3 &31.9\%  &	68.1\% \\
5 & 32.9\%  &	67.1\% \\
10 &35.0\%  &	65.0\%  \\
    \hline
    \end{tabular}
    \caption{Audio–text attention allocation ratios within demonstrations for the Khinalug dataset using Phi4, averaged over all attention heads and layers. The first column indicates the number of in-context samples.}
    \label{tab:attn_ratio_audio_text}
\end{table}

\begin{figure*}
    \centering
    \includegraphics[width=1\linewidth]{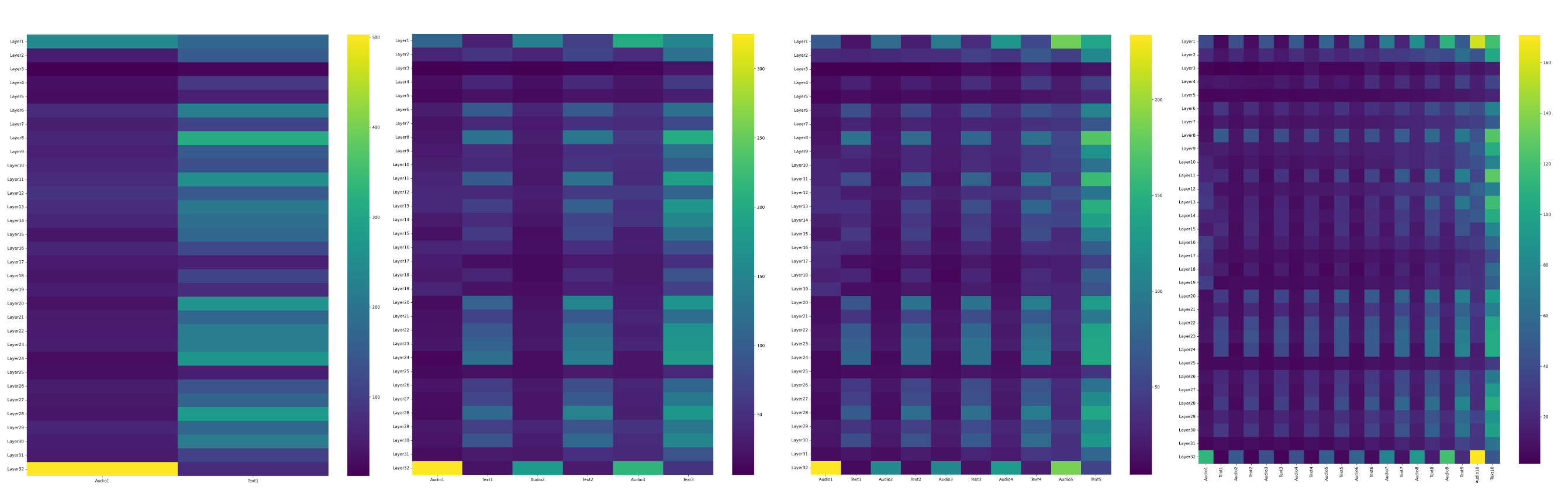}
    \caption{Layer-wise attention allocation across demonstrations of 5 and 10 samples using Phi4, averaged over all attention heads on the Khinalug dataset.}
    \label{fig:layer_wise_attn}
\end{figure*}

\textbf{Layer-dependent modality allocation:} Beyond the global allocation averaged across layers, we find a clear layer-dependent allocation to audio and text samples. Figure~\ref{fig:layer_wise_attn} shows a consistent pattern across different numbers of ICL samples: the first and last layers attend more to audio, while other layers attend more to text. One interpretation is that early layers emphasize acoustic information integration, middle layers focus on symbolic or semantic cues from text, and deeper layers re-attend to audio to support final token prediction \cite{jawahar-etal-2019-bert, de-vries-etal-2020-whats, pasad2021layer}. We note that this is a hypothesis supported by the observed attention pattern, not a guaranteed mechanism. Nevertheless, the consistency of the pattern across different sample sizes suggests that it is a stable property of the model.

Understanding layer-wise modality preferences has practical implications for interpretability and model design \cite{choi24b_interspeech, lee-etal-2025-multimodal, liu-niehues-2025-middle}. More broadly, our findings indicate that a speech LLM can organize its computation hierarchically in a modality-aware manner, offering insights for future work on improving multimodal representation learning and robustness in speech LLMs \cite{wu-etal-2025-unveiling-multimodal, basilehead, jin-etal-2025-multimodal}.

\begin{table*}[]
    \centering
    \small
    \begin{tabular}{c|cccc|cccc} \hline
          & \multicolumn{4}{|c|}{Phi4}  & \multicolumn{2}{c}{Qwen3-Omni} \\ \hline
        Language & ASR & MICL & XFT MICL &  TFT MICL  & ASR & MICL \\ \hline
         Khinalug & 138.6 & 127.6  & 111.2 & 127.6  & 108.8 & 532.4 \\
         Kichwa & 143.7 & 170.0   & 105.2 &  86.2 & 113.7 & 165.0 \\
         Mboshi & 117.8 & 144.9  & 106.9 & 38.6 & 104.2 & 112.9 \\
         \hline
         
    \end{tabular}
    \caption{Experimental results for prompting Phi4 to generate transcriptions. ASR and ICL denote task types refereed to Section~\ref{sec:prompt_design}, XFT indicates cross-lingual fine-tuned models, and TFT indicates target-language fine-tuned models with the in-context samples using the supervised data of the language.}
    \label{tab:llm_transcription}
\end{table*}

% \begin{table*}[h]
%     \centering
%     \small
%     \begin{tabular}{c|c|cccc|ccc|c} \hline
%      & &  \multicolumn{4}{c|}{Multimodal LLM} & \multicolumn{3}{|c|}{Text-based LLM} & \\ \hline
%           &  Acoustic & Phi4 ASR-FT& Phi4 XFT & Phi4 TFT  & Qwen3-Omni & Ngram-LM & Trans-LM & Llama &  Oracle \\ \hline
%          Khinalug & 42.1 & 41.5 & 41.0 & 40.8 &  40.7 & 39.6 & 41.6 & 41.9 & 36.5 \\
%          Kichwa & 17.3  & 17.4 & 17.1 & 16.6  & 17.2 & 17.7 & 18.9 & 17.0 &12.4\\
%          Mboshi & 31.4  & 29.9 & 29.6 & 28.6 &  30.0& 30.6 & 30.9 & 30.2 & 22.1 \\ 
%          \hline
%     \end{tabular}
%     \caption{Experimental results for hypothesis selection with MICL. Acoustic indicates the acoustic model output. Oracle indicates selecting the best hypothesis by the lowest WER against the ground truth. The rest indicates selecting the hypothesis with the lowest perplexity under the language model referred to Section~\ref{sec:hypo_select_approach}. XFT and TFT denote cross-lingual fine-tuning and target-language fine-tuning with Phi4.}
%     \label{tab:hypo_select_results}
% \end{table*}

\begin{table}[t]
\centering
\small
\setlength{\tabcolsep}{6pt}
\begin{tabular}{lccc}
\toprule
\textbf{Model} & \textbf{Khinalug} & \textbf{Kichwa} & \textbf{Mboshi} \\
\midrule
Acoustic    & 42.1 & 17.3 & 31.4 \\ 
\midrule
\multicolumn{4}{l}{\textit{Multimodal LLM}} \\
Phi4 ASR-FT & 41.5 & 17.4 & 29.9 \\
Phi4 XFT    & 41.0 & 17.1 & 29.6 \\
Phi4 TFT    & 40.8 & 16.6 & 28.6 \\
Qwen3-Omni  & 40.7 & 17.2 & 30.0 \\
\midrule
\multicolumn{4}{l}{\textit{Text-based LLM}} \\
Ngram-LM    & 39.6 & 17.7 & 30.6 \\
Trans-LM    & 41.6 & 18.9 & 30.9 \\
Llama       & 41.9 & 17.0 & 30.2 \\
Oracle      & 36.5 & 12.4 & 22.1 \\
\bottomrule
\end{tabular}
\caption{Experimental results for hypothesis selection with MICL. Acoustic indicates the acoustic model output. Oracle indicates selecting the best hypothesis by the lowest WER against the ground truth. The remaining methods select the hypothesis with the lowest perplexity under the language model described in Section~4.4. ASR-FT, XFT and TFT denote hypothesis selection results with the models developed from standard fine-tuning, cross-lingual fine-tuning and target-language fine-tuning, respectively.}
\label{tab:hypo_select_results}
\end{table}

\subsection{MICL enhances ASR hypothesis selection} \label{sec:hypo_selection}

\textbf{Direct ASR transcription: } Speech LLMs can be prompted to perform ASR directly. However, prior work shows that prompting-based ASR is generally weak for uncovered languages \cite{hsu2024meta, zheng2025ticl}. We verify this in our setting by prompting speech LLMs to generate transcriptions, with and without in-context demonstrations, on the datasets used in this work. As shown in Table \ref{tab:llm_transcription}, WER remain consistently high across models, indicating that MICL prompting alone is insufficient to achieve usable performance on these uncovered languages. Although TFT-MICL performs relatively better on Mboshi, which has the most supervised data among these languages, it still lags behind the acoustic models.

\textbf{MICL hypothesis selection: }To address that limitation, we combine an stronger acoustic model with a speech LLM through hypothesis selection. We use MMS as the acoustic model, because it outperforms Phi4 in fine-tuning with ASR tasks( Appendix~\ref{sec:app_ft_asr}). We select the MICL configurations for hypothesis selection based on their perplexity in the ICL evaluation. Although we do not evaluate WER for every MICL configuration, the matched results show a consistent trend that configurations with lower perplexity also yield lower WER. We therefore interpret perplexity as an empirical selection criterion rather than a direct substitute for ASR evaluation.

As the results in Table \ref{tab:hypo_select_results} show, MICL-based hypothesis selection improves over the acoustic model baseline, indicating that speech LLMs can provide useful contextual guidance when used for re-ranking. 

Inspecting the results of Phi4 fine-tuning, we find that cross-lingual instruction fine-tuning improves not only perplexity but also hypothesis selection performance. The best results are still achieved with target-language fine-tuning, highlighting the value of supervised data in the target language when available. 

Comparing to Qwen3-Omni, we conclude cross-lingual fine-tuning enables Phi4 to achieve comparable performance, even though Qwen3-Omni benefits from broad multilingual and multimodal pre-training coverage. Target-language fine-tuned Phi4 outperforms Qwen3-Omni on Kichwa and Mboshi, while the gains are smaller on Khinalug, where the amount of supervised data is the most limited. This suggests that target-language fine-tuning is most beneficial when sufficient supervised data is available.

\textbf{Text-only LM selection: }We further include several text-only re-ranking baselines to assess whether text-only models can better capture contextual information. We select off-the-shelf Llama\-3B as strong text-only LLM. We also train two language models (N-gram LM and transformer-based LM) with target-language text. The re-ranking systems are consistent with others in combining acoustic scores and LM scores, as explained in Section~\ref{sec:hypo_select_approach}.

We find multimodal speech LLMs outperform Llama across all languages in our setup. Although comparisons across LLMs are not fully controlled due to differences in architecture and pre-training data, these results suggest that multimodal conditioning is more effective for hypothesis selection than using a general-purpose text-only LLM. Besides, we observe that cross-lingual fine-tuning yields better or comparable ASR performance than a language model trained specifically for the target language but without access to target-language training data, demonstrating the effectiveness of MICL.

\textbf{Hypothesis selection VS Joint decoding: } We compare our approach with the popular joint decoding approach with an n-gram LM ( details in Appendix~\ref{sec:app_joint_decode}). Overall, joint decoding performs better than hypothesis selection in our setup. Integrating MICL in joint decoding might bring the best performance, but it is substantially more computationally expensive because joint decoding combines acoustic and language model scores at every decoding timestep, which limits practicality for large scale experiments and deployment.

\textbf{Inference-time cost:} To clarify the practical cost of the proposed approach, we conducted a preliminary measurement on the Khinalug dataset using Phi4 with a randomly selected number of paired in-context samples ranging from 1 to 10. Loading the Phi4 model requires 11 GB of GPU memory, and hypothesis selection incurs an additional 11 GB, resulting in a total memory usage of 22 GB. The average inference time for hypothesis selection is 3 seconds per item on a single NVIDIA RTX 6000 GPU. Hypothesis generation also introduces computational overhead; however, the acoustic model used in our experiments is approximately 20 times smaller than the speech LLM. This suggests that the dominant inference cost arises from the hypothesis-selection stage. Overall, these results indicate that the proposed approach is feasible for practical deployment when a sufficiently capable GPU is available.

\section{Ablation study}

\subsection{Benefits of language coverage} \label{sec:ablation_lang_coverage}

We show that cross-lingual instruction fine-tuning improves MICL performance on unseen languages. To better understand the role of multilingual coverage, we fine-tune the model using varying numbers of training languages and evaluate it on the same set of unseen languages. As shown in Table \ref{tab:XFT_num_langs}, perplexity on unseen languages consistently decreases as the number of fine-tuning languages increases, suggesting that broader language coverage leads to better cross-lingual generalization. We note that this trend may be influenced by both the language diversity and the total amount of fine-tuning data. Nevertheless, the results consistently indicate that increasing language coverage during fine-tuning is beneficial for MICL performance on unseen languages.

\begin{table}[h]
    \centering
    \small
    \begin{tabular}{cccccccc}
    \hline
         &  None & 8  & 16 & 32 & 64 & All \\ \hline
    Khinalug     & 233 & 156 & 93 & 76 & 68 & 68 \\
    Kichwa & 31 & 33 & 20 & 16 & 15 & 11 \\
    Mboshi & 75 & 49 & 36 & 33 & 27 & 28 \\
    \hline
    \end{tabular}
    \caption{Perplexity for cross-lingual fine-tuning with increasing language coverage. None indicates no fine-tuning. Numeric values denote the cumulative number of fine-tuning languages; for example, 16 extends 8 by adding 8 additional languages. The first eight languages match those covered by Phi4.}
    \label{tab:XFT_num_langs}
\end{table}

\subsection{Attention allocation reflects what model see in ICL}\label{sec:ablation_attention} 

Although we observe clear layer-dependent attention allocation patterns, attention weights alone do not guarantee that the model actually learns from the attended context. We therefore perform an intervention by replacing one in-context sample with gold information. Concretely, we replace its transcription with gold text, its audio with gold audio, or both.

As Table \ref{tab:ablation_attn_allocation} shows, replacing the text of the randomly selected sample 4 substantially reduces perplexity, and replacing both audio and text further improves performance. This aligns with the attention heatmap in Figure \ref{fig:ablation_attn_replace_gold} in Appendix~\ref{sec:app_replace_gold}, where attention shifts more strongly toward sample 4 after replacement. Together, these results suggest that attention allocation reflects actual in-context usage and that the model relies on attended samples for prediction.

Moreover, Table~\ref{tab:ablation_attn_allocation} and Figure~\ref{fig:ablation_attn_replace_gold} shows that replacing the text sample increases attention to the corresponding audio sample, suggesting that the model understands MICL task by connecting the paired text and audio. In contrast, replacing only the audio yields little improvement, possibly because text is easier for the model to exploit, or audio-only cues are harder to leverage without strong textual reference.

% TBD: resutls with text-only ICL; split audio or text into half and half to explore attention allocation; Oracle selection results with providing the gold sample.

\begin{table}[h!]
\centering
\small
\begin{tabular}{c|c|cc}
\hline

& Perplexity  & Audios & Texts\\
\hline

    No change & 68   & 35.0\%  & 65.0\%  \\
  Replace text & 9  & 43.3\%  & 56.7\%  \\
  Replace Audio & 67   & 36.1\%  & 63.9\%  \\
  Replace pair & 2   & 46.7\%  & 53.3\%  \\
 
\hline
\end{tabular}
\caption{Evaluation results on perplexity and attention allocation across audio and text when replacing one in context example with a gold sample using different replacement strategies for text audio or both.}
\label{tab:ablation_attn_allocation}
\end{table}

\section{Conclusion}

This work shows that MICL is effective at learning unseen low-resource languages by leveraging both speech and text modalities. We demonstrate that prompt-based ASR performs poorly on unseen languages, and we propose a MICL-based hypothesis selection approach that effectively combines MICL’s contextual understanding with a strong acoustic model. We further find that cross-lingual transfer learning without seeing target languages achieves performance comparable to models trained with target-language data, indicating the benefits of broader language coverage during LLM development. Finally, we reveal that MICL exhibits layer-dependent attention allocation with a preference for text samples, offering insights for developing more interpretable and effective multimodal LLMs.


\section*{Limitations}
% Required by ACL!!!

We evaluate MICL on a single ASR task. While this task is challenging, evaluating MICL on additional speech and multimodal tasks would further validate its generality. In addition, our experiments are conducted on three languages. Although the results are consistent across them, the overall language coverage remains limited. Our evaluation is restricted to two state-of-the-art open-source speech LLMs, as the proposed hypothesis selection approach requires access to model parameters. As a result, we do not assess MICL on closed-source LLMs, such as Gemini or GPT, and our findings may not directly generalize to such models.

\section*{Ethics Statement}

This work relies exclusively on publicly available datasets and models, and we do not identify direct ethical concerns associated with the proposed approach. The methods studied are general-purpose techniques for low-resource language adaptation and do not explicitly model or target language-specific or societal biases. ChatGPT was used only for grammar correction of author-written text during manuscript preparation.

% ACL encourages the authors to include an optional section dedicated to discussing the broader impacts and ethical considerations of the submission. Likewise, it may not contain any additional experiments, figures or analysis. This section should be placed after the conclusion and before references, without page breaks. Its content does not count towards the page limit.

\section*{Acknowledgments}
This work was supported by the Helmholtz Programme-oriented Funding, with project number 46.24.01, project name AI for Language Technologies. We acknowledge the HoreKa supercomputer funded by the Ministry of Science, Research and the Arts Baden-Wurttemberg and by the Federal Ministry of Education and Research.

\bibliography{custom}

\appendix

\section{Experimental setups}\label{sec:app:exp_setup}

We experiment with \textit{Phi4-multimodal-instruct}\footnote{\url{https://huggingface.co/microsoft/Phi4-multimodal-instruct}} and \textit{Qwen3-Omni-30B-A3B-Instruct}\footnote{\url{https://huggingface.co/Qwen/Qwen3-Omni-30B-A3B-Instruct}} under constraints of computation resources.

We experiment with  NVIDIA RTX 6000 Ada with 48GB for Phi4 and Nvidia RTX PRO 6000 Blackwell 96 GB for Qwen3-Omni. With hyperparameter tuning, we implement cross-lingual fine-tuning with optimizer adamw\_torch. The learning rate is 1e-5 with weight decay of 0.01. The warm-up step is 500 and the batch size is 8. The details refer to the scripts.

The hypotheses generation model is \textit{mms-300m}\footnote{\url{https://huggingface.co/facebook/mms-300m}}. We use \textit{Llama-3-8B-Instruct}\footnote{\url{https://huggingface.co/meta-llama/Meta-Llama-3-8B-Instruct}} as for text-only LLM hypothesis selection.

\section{Join Decoding and Hypothesis Selection}\label{sec:app_joint_decode}

We propose a hypothesis selection approach using MICL to improve ASR performance with an acoustic model in this work. Another way to improve ASR performance of an acoustic model is to perform joint decoding with a language model. We compare these two approaches and report results in Table~\ref{tab:joint_decode}. Joint decoding consistently outperforms hypothesis selection. It also exceeds the oracle results on Khinalug, which is the language with the smallest amount of labeled data in our setup. As the amount of labeled data increases, the performance gap between joint decoding and hypothesis selection narrows. Overall, joint decoding is the stronger approach in low resource settings, and the results suggest further gains may be possible by incorporating large language models into joint decoding.

\begin{table}[h]
    \centering
    \small
    \begin{tabular}{cccc} \hline
    & Joint decode & Hypo select & Hypo oracle \\ \hline
    Khinalig     &  34.2 & 40.8 & 36.5 \\ 
    Kichwa     & 15.4 & 16.6 & 12.4 \\
    Mboshi & 27.3 & 28.6 & 22.1 \\
    \hline
    \end{tabular}
    \caption{Comparison between joint decoding with an n-gram langauge model and hypothesis selection in ASR evaluted with WER.}
    \label{tab:joint_decode}
\end{table}

\section{Fine-tune with Target Language on ASR Task}\label{sec:app_ft_asr}

To support our statement that the acoustic model is better than speech LLMs in building ASR models for unseen languages, we fine-tune speech LLMS following the standard ASR prompt. Comparing Table~\ref{tab:app:ft_asr_results} and Table~\ref{tab:hypo_select_results}, we can see the acoustic model shows clearly better performance than the speech LLMs, indicating the advantages of self-supervised learning in low-resource languages.

\begin{table}[h]
    \small
    \centering
    \begin{tabular}{ccc} \hline
         &  MMS & Phi4 \\ \hline
    Khinalug    & 42.1 & 77.9 \\
    Kichwa & 17.3 & 40.3 \\
    Mboshi & 31.4 & 42.3 \\
    \hline
    \end{tabular}
    \caption{Results of fine-tuning with target language on ASR task.}
    \label{tab:app:ft_asr_results}
\end{table}

\section{Prompt Design} \label{sec:app_prompt}
Here are the prompt examples for T-ICL, ICL and MICL with Phi4. The ones for Qwen3-omni and the implementation details are available in the github repo.

\begin{tcolorbox}[
  colback=gray!5,
  colframe=gray!50,
  boxrule=0.5pt,
  fontupper=\ttfamily\small,
  breakable
]

<|system|>You are an expert at learning the language from provided samples.

<|user|>Here are sample texts.

Transcription: text\_1

Transcription: text\_2

......

Transcription: text\_n

<|end|><|assistant|>
\end{tcolorbox}

\begin{tcolorbox}[
  colback=gray!5,
  colframe=gray!50,
  boxrule=0.5pt,
  fontupper=\ttfamily\small,
  breakable
]

<|system|>You are an expert at learning the language from provided samples. You are also a helpful and accurate AI model that transcribes audio clips into written text in that language. 

<|user|>Here are sample texts.

Transcription: text\_1

Transcription: text\_2

......

Transcription: text\_n
<|audio\_n+1> Transcribe the audio clip into text and output only the transcription<|end|><|assistant|>

\end{tcolorbox}

\begin{tcolorbox}[
  colback=gray!5,
  colframe=gray!50,
  boxrule=0.5pt,
  fontupper=\ttfamily\small,
  breakable
]
<|system|>You are an expert at learning the language from provided samples. You are also a helpful and accurate AI model that transcribes audio clips into written text in that language.

<|user|>Here are sample pairs of audio and text.

<|audio\_1> Transcription: text\_1

<|audio\_2> Transcription: text\_2

......

<|audio\_n> Transcription: text\_n

<|audio\_n+1> Transcribe the audio clip into text and output only the transcription<|end|><|assistant|>
\end{tcolorbox}

\section{Representation Length Analysis}\label{sec_app_audio_text_repre_lengths}

Experiment with Phi4 and Khinalug dataset, we count the representation length of audio and text samples to support attention allocation analysis. As Table~\ref{tab:audio_text_repre_lengths} shows, the audio representation lengths is around three times longer than that of text, on average of the first 20 test data with 10 in-context samples

\begin{table}[h]
    \centering
    \small
    \begin{tabular}{ccc} \hline
    Text data     & Audios & Texts   \\ \hline
    1  & 302  & 126 \\
2  & 576  & 231 \\
3  & 331  & 153 \\
4  & 1794 & 512 \\
5  & 857  & 289 \\
6  & 1023 & 329 \\
7  & 1296 & 349 \\
8  & 1664 & 500 \\
9  & 711  & 274 \\
10 & 412  & 171 \\
11 & 459  & 183 \\
12 & 298  & 105 \\
13 & 1508 & 431 \\
14 & 1232 & 355 \\
15 & 704  & 256 \\
16 & 864  & 251 \\
17 & 312  & 174 \\
18 & 537  & 191 \\
19 & 446  & 169 \\
20 & 1915 & 571 \\
\hline
Sum & 17241 & 5620 \\ \hline
    \end{tabular}
    \caption{Representation lengths for audio and text samples}
    \label{tab:audio_text_repre_lengths}
\end{table}

\section{Attention Allocation Results}\label{sec:app_attn_allocation}

As described in Section~\ref{sec:MICL_interpret}, in this section, we show the full attention allocation results  in Table~\ref{tab:attn_allocation}. Overall, most attention goes into prompt, which is reasonable as the LLM is fine-tuned for instruction following. The target audio gets more attention than the demonstrations when the number of sample is one. However, there is more attention goes into demonstration samples with the increase of the number of sample, and we assume this might because of the increasing representation length of the samples.

\begin{table*}[h!]
\centering
\small
\begin{tabular}{c|cccccc}
\hline

& \#samples & Prompt & Audios& Texts& Target audio& Task \\
\hline
Phi4 &  1  & 76.1\% & 3.3\% & 5.6\% & 9.7\% & 5.3\% \\
 & 3  & 71.4\% & 6.3\% & 9.7\% & 7.1\% & 5.6\% \\
 & 5  & 69.3\% & 7.7\% & 11.1\% & 5.9\% & 5.9\% \\
 & 10 & 66.9\% & 9.3\% & 12.5\% & 4.7\% & 6.6\%  \\
\hline
Phi4 XFT & 1  & 74.4\% & 2.8\% & 6.3\% & 11.0\% & 5.6 \\
 & 3  & 69.9\% & 4.6\% & 9.8\% & 9.9\% & 5.8\%  \\
 & 5  & 68.2\% & 5.4\% & 11.0\% & 9.3\% & 6.1\%  \\
 & 10 & 65.9\% & 6.6\% & 12.2\% & 8.7\% & 6.7\%  \\

& 10 (Replace text 4) & 63.9\%  & 9.7\%  & 12.7\%  & 7.0\%  & 6.7\%  \\
& 10 (Replace audio 4)& 64.7\%  & 7.5\%  & 13.3\%  & 7.8\%  & 6.7\% \\
& 10 (Replace pair 4) & 63.0\%  & 11.1\%  & 12.7\%  & 6.5\%  & 6.7\%    \\

\hline
\end{tabular}
\caption{Attention allocation for different samples with Phi4 and cross-lingual FT LoRA settings.}
\label{tab:attn_allocation}
\end{table*}

\section{Attention Allocation for Replacing with Gold Samples} \label{sec:app_replace_gold}
\begin{figure*}[h]
    \centering
    \includegraphics[width=1\linewidth]{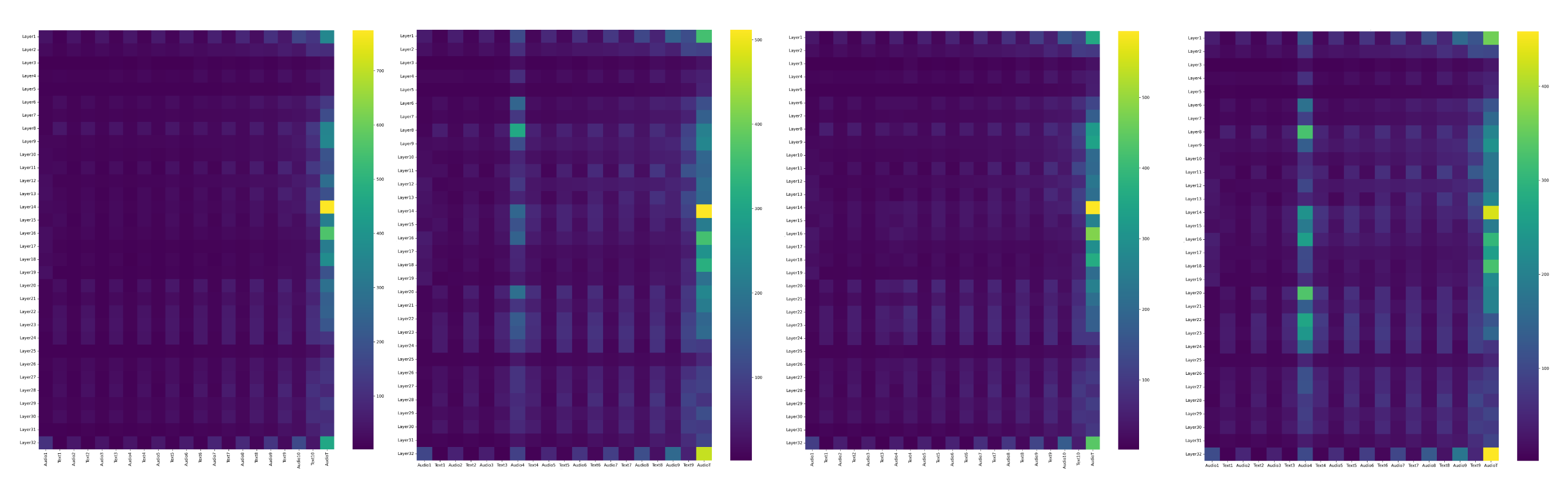}
    \caption{Layer-wise attention allocation under different intervention settings. We show attention allocation with no modification (left 1), with gold text replacing sample 4 (left 2), with gold audio replacing sample 4 (right 2), and with both gold text and audio replacing sample 4 (right 1). We include the attentions on target audio for reference purposes.}
    \label{fig:ablation_attn_replace_gold}
\end{figure*}
To study whether attention allocation is actually related to model learning, we perform intervention experiments by replacing the fourth sample with gold information. Figure \ref{fig:ablation_attn_replace_gold} shows the attention heatmaps with and without replacement. Unlike Figure~\ref{fig:layer_wise_attn}, which includes attention only on the in-context samples, here we also include attention to the target audio for reference. 

Overall, most attention goes into prompt, which is reasonable as the LLM is fine-tuned for instruction following. With the increase of the number of samples, there is more attention goes into demonstration, but we assume this might because of the increasing representation length of the samples. 

We also observe that replacing the text or the text–audio pair with gold information significantly increases the corresponding attention, supporting the analysis in Section~\ref{sec:MICL_interpret} and Section~\ref{sec:ablation_attention}. Notably, attention to the gold information is comparable to or higher than that to the target audio, indicating a strong correlation between attention weights and model learning.

\section{Sample Selection Strategy} \label{sec:app_sample_selection}

We evaluate four sample selection strategies: random, text-based, audio-based, and text–audio–combined. We use the same acoustic model of the hypotheses generation to generate the transcript of the target audio, then search relevant samples based on the prediction. The implementation details are available in the scripts. As shown in Table~\ref{tab:ppl_phi_sample_strategy}, all selection strategies outperform random sampling, indicating their effectiveness. Among them, text-based and text–audio–combined selection achieve strong overall performance, with text-based selection performing particularly well under longer context lengths. Based on these results, we adopt text-based sample selection in this work.

\begin{table*}[h]
\centering
\small
\setlength{\tabcolsep}{4pt}
\begin{tabular}{cccccccccccc}
\hline
Language & Type & Task  & 0 & 1 & 2 & 3 & 5 & 10 & 25 & 50 & 100 \\
\hline

Khinalug & Random  & T-ICL  & 7392 & 1573 & 1156 & 967 & 861 & 686 & 438 & 345 & 287 \\
 &  & ICL       & 1445 & 550 & 445 & 418 & 396 & 449 & 324 & 300 & 298 \\
 & & MICL    & 1508 & 580 & 482 & 466 & 532 & 658 & 552 & 534 & 447 \\
 \cline{2-12}
 
 &  Text-based  & T-ICL     & 7435 & 1347 & 955 & 779 & 328 & 248 & 193 & 166 & 175 \\
 &   & ICL     & 1445 & 909 & 606 & 1778 & 189 & 154 & 143 & 147 & 158 \\
 &  & MICL      & 1508 & 485 & 318 & 545 & 261 & 230 & 273 & 284 & 241 \\
  \cline{2-12}
  
  &  Audio-based & T-ICL     & 7392 & 1578 & 1013 & 857 & 656 & 525 & 344 & 307 & 25 \\
 &   & ICL      & 1497 & 880 & 429 & 378 & 342 & 343 & 285 & 288 & 262 \\
 &  &MICL     & 1580 & 896 & 513 & 443 & 431 & 464 & 476 & 443 & 386 \\
  \cline{2-12}
  
  &  Text-Audio-combined &  T-ICL     & 7392 & 1037 & 663 & 553 & 437 & 281 & 223 & 200 & 189 \\
 &  & ICL      & 1497 & 326 & 285 & 253 & 215 & 190 & 190 & 188 & 200 \\
 &   & MICL      & 1580 & 341 & 334 & 298 & 271 & 306 & 304 & 346 & 276 \\

\hline

Kichwa & Random &  T-ICL & 35157 & 5459 & 2119 & 2150 & 2181 & 2390 & 2780 & 1025 & 1267 \\
  &   & ICL   & 132 & 72 & 56 & 58 & 48 & 42 & 44 & 50 & 55 \\
 &   & MICL    & 136 & 85 & 57 & 60 & 66 & 60 & 67 & 69 & 230  \\
\cline{2-12}
   
 &  Text-based & T-ICL &  35157 & 3141 & 1568 & 1125 & 836 & 424 & 534 & 266 & 253 \\
&   & ICL     & 132 & 65 & 78 & 70 & 74 & 39 & 30 & 33 & 26 \\
 &   & MICL     & 136 & 64 & 58 & 43 & 48 & 31 & 40 & 58 & 69 \\
   \cline{2-12}
   
  &  Audio-based & T-ICL &  35157 & 3157 & 1260 & 1018 & 880 & 1048 & 1302 & 857 & 1003 \\
  &   & ICL   & 132 & 64 & 47 & 45 & 42 & 38 & 37 & 38 & 39 \\
 &   & MICL     & 136 & 85 & 58 & 54 & 45 & 57 & 95 & 98 & 191 \\
  \cline{2-12}
 
  &  Text-Audio-combined & T-ICL &  35157 & 4297 & 1353 & 982 & 872 & 482 & 502 & 544 & 532 \\
  &   & ICL     & 132 & 68 & 70 & 61 & 49 & 40 & 39 & 33 & 36 \\
 &   & MICL    & 136 & 91 & 64 & 46 & 38 & 42 & 81 & 69 & 100 \\

\hline

\end{tabular}
\caption{Perplexity results of Phi4 with different sample selection strategies.}
\label{tab:ppl_phi_sample_strategy}
\end{table*}

% \begin{table}[]
%     \centering
%     \small
%     \begin{tabular}{ccccc} \hline
%          & High & Middle & Low & Few Shot\\ \hline
%     Phi4-mm     & \\
%     Whisper & \\
%     ICL & \\
%     ICL Oracle & \\
%     % Hypo select ICLL & \\
%     \hline
%     \end{tabular}
%     \caption{Caption}
%     \label{tab:placeholder}
% \end{table}

\section{Phi4 Fine-tuning Strategies} \label{sec:app:ft_startegies}
This work fine-tune Phi4 with the activate parameters in audio encoder, projector, and lora adapter and frozen parameters of all the rest. This section quantity the effectiveness and computational cost of different fine-tuning modules. As Table \ref{tab:app_ft_module} shows, there is no clear different between fine-tuning only with LORA modules in decoder or with additional modules.

\begin{table*}
\centering
\small
\setlength{\tabcolsep}{4pt}
\begin{tabular}{ccccccccccc}
\hline
Module & \# Params & 0 & 1 & 2 & 3 & 5 & 10 & 25 & 50 & 100 \\ \hline

LORA & 461M & 609 & 83 & 55 & 43 & 36 & 28 & 26 & 30 & 30 \\ 
LORA + Projector & 487M & 610 & 83 & 55 & 43 & 36 & 28 & 26 & 31 & 29 \\
LORA + Projector + Audio Encoder & 928M & 568 & 77 & 51 & 40 & 33 & 25 & 23 & 27 & 26 \\ \hline
\end{tabular}
\caption{Experimental results with different fine-tuning strategies with Phi4}
\label{tab:app_ft_module}
\end{table*}

\section{Fine-tuning Number of In-context Samples}\label{sec:app_ft_num_sample}

We evaluate cross-lingual fine-tuning under different sample-count settings. As shown in Table~\ref{tab:ft_num_sample}, randomly selecting samples and increasing the number of samples generally lead to better performance.

\begin{table*}[]
\centering
\small
\begin{tabular}{ccccccccccc} \hline

& \#Samples & 0 & 1 & 2 & 3 & 5 & 10 & 25 & 50 & 100 \\ \hline
Khinalug & 1-5     & 778  & 183 & 231 & 230 & 76 & 72 & 82 & 104 & 112 \\
& 1-10    & 838  & 188 & 193 & 210 & 73 & 68 & 76 & 95  & 98  \\
& Fix 5  & 1089 & 213 & 256 & 266 & 70 & 66 & 76 & 96  & 97  \\
& Fix 10 & 1034 & 200 & 218 & 241 & 64 & 59 & 68 & 84  & 87  \\
\hline
% \multirow{4}{*}{Kchwa}
Kichwa &  1-5     & 63   & 19  & 16  & 14  & 13 & 10 & 9  & 10  & 14  \\
& 1-10    & 65   & 22  & 19  & 14  & 13 & 11 & 10 & 10  & 14  \\
& Fix 5  & 77   & 23  & 18  & 14  & 12 & 10 & 9  & 9   & 13  \\
&Fix 10 & 85   & 25  & 19  & 14  & 12 & 9  & 8  & 8   & 10  \\
\hline
\end{tabular}
\caption{Results under different number of samples during cross-lingual fine-tuning for Khinalug and Kchwa. 1-5 indicates the sample number is randomly selected between 1 and 5 for each instance. and fix 5 indicates a fixed number of 5 samples per instance}
\label{tab:ft_num_sample}
\end{table*}

\end{document}